# Advanced phase retrieval: maximum likelihood technique with sparse regularization of phase and amplitude

A. Migukin[*], V. Katkovnik and J. Astola

Department of Signal Processing, Tampere University of Technology, P.O. Box 553, FI-33101, Tampere, Finland

**Abstract:**
Sparse modeling is one of the efficient techniques for imaging that allows recovering lost information. In this paper, we present a novel iterative phase-retrieval algorithm using a sparse representation of the object amplitude and phase. The algorithm is derived in terms of a constrained maximum likelihood, where the wave field reconstruction is performed using a number of noisy intensity-only observations with a zero-mean additive Gaussian noise. The developed algorithm enables the optimal solution for the object wave field reconstruction. Our goal is an improvement of the reconstruction quality with respect to the conventional algorithms. Sparse regularization results in advanced reconstruction accuracy, and numerical simulations demonstrate significant enhancement of imaging.

**Introduction**

The conventional sensors detect only the intensity of the light, but the phase is systematically lost in measurements. Phase retrieval is a problem of the phase recovering using a number of intensity observations and some prior on the object wave field. The phase carries important information about an object shape what is necessary for a 3D object imaging and exploited in many areas such as microscopy, astronomy, etc. Moreover, phase-retrieval techniques are often simpler, cheaper and more robust comparing with interferometric ones.

In 1982 Fienup introduced some, for now classical, iterative phase-retrieval algorithms [1]: error-reduction, gradient search and input-output methods. Many phase-retrieval methods are developed based on this pioneer work: the estimated magnitudes at the measurement planes are iteratively replaced by ones obtained from the intensity observations.

---

[*] This work was supported by the Academy of Finland: project no. 213462, 2006-2011 (Finnish Programme for Centres of Excellence in Research) and project no. 138207, 2011-2014; and the postgraduate work of Artem Migukin is funded by Tampere Doctoral Programme in Information Science and Engineering (TISE).

We are looking for the optimal wave field reconstruction from a number of intensity observations, and the reconstruction problem is formulated in terms of a variational constrained maximum likelihood (ML) approach. The spatial image resolution of the conventional phase-retrieval techniques is limited due to diffraction, what can be one of the main sources of artifacts and image degradations. In order to enhance the imaging quality and recover lost information, in this work we use the novel compressive sensing technique for the variational image reconstruction originated in [2]. The object wave field distribution is assumed to be sparse, and its amplitude and phase are separately analyzed and decomposed using very specific basis functions named BM3D-frames [3]. The proposed phase-retrieval algorithm is derived as a solution of the ML optimization problem using the BM3D-frame based sparse approximation of the object amplitude and phase distributions.

**Propagation model**

We consider a multi-plane wave field reconstruction scenario: a planar laser beam illuminates an object, and the result of the wave field propagation is detected on a sensor at different distances $z_r$ from the object at various measurement (sensor) planes. Here $z_r = z_1+(r-1)\cdot\Delta_z$, $r=1,\ldots K$, where $z_1$ is the distance from the object to the first measurement plane, $\Delta_z$ is the distance between the measurement planes, and $K$ is a number of these planes. We assume that the wave field distributions at the object and sensor planes are pixel-wise invariant. In such a discrete-to-discrete model, the forward wave field propagation from the object to the $r$-th sensor plane is defined in the vector-matrix form as follows:

$$\mathbf{u}_r = \mathbf{A}_r \cdot \mathbf{u}_0, \ r=1,\ldots K, \qquad (1)$$

where $\mathbf{u}_0$ and $\mathbf{u}_r$ are $\mathbb{C}^n$ vectors, constructed by columns concatenating 2D discrete complex-valued wave field distributions ($N \times M$ matrices) at the object and sensor planes, respectively. $\mathbf{A}_r \in \mathbb{C}^{n \times n}$ is a discrete forward propagation operator, $n=N\cdot M$. We consider the paraxial approximation of the wave field propagation defined by the Rayleigh-Sommerfield integral. Depending of the used discretization of this integral, the operators $\mathbf{A}_r$ in (1) can be, for instance, the angular spectrum decomposition or the discrete diffraction transform in the matrix (M-DDT, [4]) or the Fourier transform domains (F-DDT, [5]). In our numerical experiments, we use F-DDT models enabling the exact pixel-to-pixel mapping $\mathbf{u}_0$ to $\mathbf{u}_r$.

According to the used vector-matrix notation, the observation model with the additive zero-mean Gaussian noise at the sensor planes, $\varepsilon_r[k] \sim N(0,\sigma_r^2)$, takes the form:

$$\mathbf{o}_r = |\mathbf{u}_r|^2 + \varepsilon_r, \ r=1,\ldots K \qquad (2)$$

Let us assume that the object amplitude $\mathbf{a}_0 \in \mathbb{R}^n$ and phase $\boldsymbol{\varphi}_0 \in \mathbb{R}^n$ can be separately approximated by small numbers of basic functions with coefficients $\boldsymbol{\theta}_a$ for the amplitude and coefficients $\boldsymbol{\theta}_\varphi$ for the phase. These basic functions are collected in the matrices $\boldsymbol{\Psi}_a$ and $\boldsymbol{\Psi}_\varphi$ for the amplitude and the phase, respectively. The amplitude and the phase are reconstructed from the noisy intensity data $\mathbf{o}_r$.

**Decoupled augmented Lagrangain (D-AL) algorithm**

According to the maximum likelihood approach, the reconstruction of the object wave field is performed by minimization of the criterion:

$$L = \sum_{r=1}^{K} \frac{1}{2\sigma_r^2} \| \mathbf{o}_r - |\mathbf{u}_r|^2 \|_2^2 + \tau_a \cdot \| \boldsymbol{\theta}_a \|_{l_p} + \tau_\varphi \cdot \| \boldsymbol{\theta}_\varphi \|_{l_p}, \qquad (3)$$

subject to the following constraints: first of all the forward propagation models (1), and the constraints for the sparse modeling for the object amplitude and phase given as

$$\mathbf{a}_0 = \boldsymbol{\Psi}_a \cdot \boldsymbol{\theta}_a, \qquad \boldsymbol{\theta}_a = \boldsymbol{\Phi}_a \cdot \mathbf{a}_0, \qquad (4)$$

$$\boldsymbol{\varphi}_0 = \boldsymbol{\Psi}_\varphi \cdot \boldsymbol{\theta}_\varphi, \qquad \boldsymbol{\theta}_\varphi = \boldsymbol{\Phi}_\varphi \cdot \boldsymbol{\varphi}_0. \qquad (5)$$

The quadratic fidelity term in (3) appears due to our assumption that the observation noise is Gaussian. The next two terms define the sparse regularization in the spectral domain, where the positive parameters $\tau_a$ and $\tau_\varphi$ define a balance between the fit of observations, the smoothness of the wave field reconstruction and the complexity of the solution.

The first equations in (4) and (5) give the restrictions for the amplitude and the phase in the synthesis form and the second ones - in the analysis form [2, 3]. Note that the priori unknown basic functions for the amplitude and the phase are selected from the synthesis $\boldsymbol{\Psi}_a$, $\boldsymbol{\Psi}_\varphi$ and analysis $\boldsymbol{\Phi}_a$, $\boldsymbol{\Phi}_\varphi$ matrices defining the redundant sets of the basic functions. The vectors $\boldsymbol{\theta}_a$, $\boldsymbol{\theta}_\varphi \in \mathbb{R}^m$ are considered as *spectra* for parametric approximations of the amplitude and the phase. Conventionally, the sparsity of these approximations is characterized by a number of non-zero components of the spectra vectors $\boldsymbol{\theta}$ ($l_0$-norm, $||\boldsymbol{\theta}||_0$, of the vector $\boldsymbol{\theta}$) or by the sum of the absolute values of the vector elements ($l_1$-norm, $||\boldsymbol{\theta}||_1 = \sum_s |\boldsymbol{\theta}_s|$, of the vector $\boldsymbol{\theta}$). In this work, the $l_1$-norm is used based on the results stating that the solutions obtained for the $l_0$- or $l_1$-norms are close to each other [6].

We use the augmented Lagrangian approach in order to reduce the constrained optimization for (3)-(5) to the unconstrained one. Furthermore, following the decoupling technique originated in [2, 3] instead of optimization of a single criterion we use the alternating optimization of two criteria $L_1$ and $L_2$:

$$L_1=\sum_{r=1}^{K}\frac{1}{\sigma_r^2}[\frac{1}{2}\|\mathbf{o}_r-|\mathbf{u}_r|^2\|_2^2+\frac{1}{\gamma_r}\|\mathbf{u}_r-\mathbf{A}_r\cdot\mathbf{u}_0\|_2^2+\frac{2}{\gamma_r}\text{Re}\{\mathbf{\Lambda}_r^H(\mathbf{u}_r-\mathbf{A}_r\cdot\mathbf{u}_0)\}]+\frac{1}{\xi}\|\mathbf{u}_0-\mathbf{v}_0\|_2^2 \quad (6)$$

$$L_2=\tau_a\cdot\|\mathbf{\theta}_a\|_{l_p}+\tau_\varphi\cdot\|\mathbf{\theta}_\varphi\|_{l_p}+\frac{1}{2\gamma_a}\|\mathbf{\theta}_a-\mathbf{\Phi}_a\cdot\mathbf{a}_0\|_2^2+\frac{1}{2\gamma_\varphi}\|\mathbf{\theta}_\varphi-\mathbf{\Phi}_r\cdot\mathbf{\varphi}_0\|_2^2 \quad (7)$$

The criterion $L_1$ in (6) is formed from the fidelity term and the forward propagation constraints similar to [7]. The variable $\mathbf{v}_0$ in the latter summand serves as a splitting variable separating optimization of $L_1$ and $L_2$. This variable is an estimate of the object distribution $\mathbf{u}_0$. It is calculated as $\mathbf{v}_0=\mathbf{\Psi}_a\cdot\mathbf{\theta}_a\circ\exp(j\cdot\mathbf{\Psi}_\varphi\cdot\mathbf{\theta}_\varphi)$, where '∘' stands for the element-wise multiplication of two vectors. $\xi$, $\gamma_r$, $\gamma_a$ and $\gamma_\varphi$ are positive parameters controlling "weight" of these penalty terms. $(\cdot)^H$ stands for the Hermitian conjugate.

A typical Lagrangian based optimization assumes: minimization of $L_2$ with respect to $\mathbf{\theta}_a$, $\mathbf{\theta}_\varphi$; minimization of $L_1$ on $\mathbf{u}_0$, $\{\mathbf{u}_r\}$, and maximization of $L_1$ with respect to the vectors of the Lagrange multipliers $\mathbf{\Lambda}_r\in\mathbb{C}^n$. The splitting variable $\mathbf{v}_0$ separates minimizations on $\mathbf{u}_0$ and on $\mathbf{\theta}_a$, $\mathbf{\theta}_\varphi$. The solutions obtained for optimization of $L_1$ and $L_2$ results in the following iterative algorithm, similar to [2]:

Initialize $\mathbf{u}_0^0$, $\{\mathbf{\Lambda}_r^0\}$ and calculate transform matrices $\mathbf{\Psi}_a$, $\mathbf{\Psi}_\varphi$, $\mathbf{\Phi}_a$, $\mathbf{\Phi}_\varphi$ for $t = 0$

Repeat for $t =1, 2\ldots$

$$\mathbf{\theta}_a^t=\Im h_{\tau_a\gamma_a}(\mathbf{\Phi}_a\mathbf{a}_0^t),\quad \mathbf{\theta}_\varphi^t=\Im h_{\tau_\varphi\gamma_\varphi}(\mathbf{\Phi}_\varphi\mathbf{\varphi}_0^t)$$

$$\mathbf{v}_0^t=\mathbf{\Psi}_a\mathbf{\theta}_a^t\circ\exp(j\mathbf{\Psi}_\varphi\mathbf{\theta}_\varphi^t),\quad \mathbf{u}_r^{t+1/2}=\mathbf{A}_r\mathbf{v}_0^t$$

$$\mathbf{u}_r^{t+1}[k]=\mathcal{G}(\mathbf{o}_r[k],\mathbf{u}_r^{t+1/2}[k],\mathbf{\Lambda}_r^t[k]),\quad r=1,\ldots K \quad (8)$$

$$\mathbf{\Lambda}_r^{t+1}=\mathbf{\Lambda}_r^t+\alpha_r\cdot(\mathbf{u}_r^{t+1}-\mathbf{u}_r^{t+1/2}),\quad r=1,\ldots K$$

$$\mathbf{u}_0^{t+1}=(\sum_{r=1}^{K}\frac{1}{\sigma_r^2\gamma_r}\mathbf{A}_r^H\mathbf{A}_r+\frac{1}{\xi}\mathbf{I})^{-1}\times(\sum_{r=1}^{K}\frac{1}{\sigma_r^2\gamma_r}\mathbf{A}_r^H(\mathbf{u}_r^{t+1}+\mathbf{\Lambda}_r^t)+\frac{\mathbf{v}_0^t}{\xi})$$

End on $t$

The initialization for $t=0$ concerns the object distribution $\mathbf{u}_0^0$, Lagrangian multipliers (usually $\{\mathbf{\Lambda}_r^0[k]\}=0$) and the BM3D-frames (matrices $\mathbf{\Psi}_a$, $\mathbf{\Psi}_\varphi$, $\mathbf{\Phi}_a$, $\mathbf{\Phi}_\varphi$) used for the synthesis and analysis for both the object amplitude and phase. The updates of $\{\mathbf{u}_r^{t+1}\}$ are obtained by minimization of $L_1$ with respect to $\{\mathbf{u}_r\}$, what results in a statistically optimal fitting of $\{\mathbf{u}_r^{t+½}\}$ to the observations $\mathbf{o}_r$. It is computed by the operator $\mathcal{G}$ defined similar to [7]. The soft thresholding operator obtained by minimization of $L_2$ with respect to $\mathbf{\theta}_a$, $\mathbf{\theta}_\varphi$ is calculated as

$$\mathbf{\theta}=\Im h_\tau(\mathbf{u})=sign(\mathbf{u})\circ(|\mathbf{u}|-\tau)_+ \quad (9)$$

We name the proposed algorithm *Decoupled Augmented Lagrangian* (D-AL) *algorithm*. The main difference of this algorithm with respect to the AL algorithm in [7] is that in the D-AL algorithm the phase and amplitude estimates at the object plane are filtered using the BM3D-frame sparse representations, what result in a significant imaging enhancement.

**Numerical experiments**

In simulation experiments, we compare three phase-retrieval algorithms: AL from [7], the forward-backward (FB) variation of SBMIR from [8] and the proposed D-AL algorithm (8). Here we consider a phase-only object distributions given for $\mathbf{u}_0=1\cdot\exp(j\cdot\pi\cdot(\mathbf{w}-½))$, where $\mathbf{w}$ is a binary *chessboard* test-image (128×128). The wave fields $\mathbf{u}_r$ and $\mathbf{u}_0$ are pixilated with square pixels $\Delta\times\Delta$, $\Delta$= 6.7µm, and 100% fill factors. The results are given for noisy observations with $\sigma_r=\sigma=0.05$ for all $r$, and the following setup parameters: wavelength $\lambda$=532nm, $\Delta_z$=2mm, $z_1$=2·$z_f$, $z_f$ is "in-focus" distance. The number of the observation planes $K$=5. A good initial estimate $\mathbf{u}_0^0$ is important for BM3D filtering. The object initialization is calculated by AL with 50 iterations, then the object wave field reconstruction is performed using another 50 iterations of the D-AL algorithm.

In Fig. 1, the reconstructed object amplitude and phase are shown after 100 iterations of the considered phase-retrieval algorithms. The visual advantage of the D-AL algorithm is obvious. The reconstruction accuracy is characterized by root-mean-square error (RMSE) calculated for the whole image. The cross-sections of these reconstructions are illustrated in Fig. 2. The best performance of the D-AL algorithm is clear. These D-AL reconstructions are very close to the true object phase and amplitude, while the AL and SBMIR reconstructions are blurred and show large reconstruction errors.

Numerical experiments demonstrate a significant better reconstruction quality of D-AL: in RMSE values, it is approximately ten times better with respect to AL and more for SBMIR.

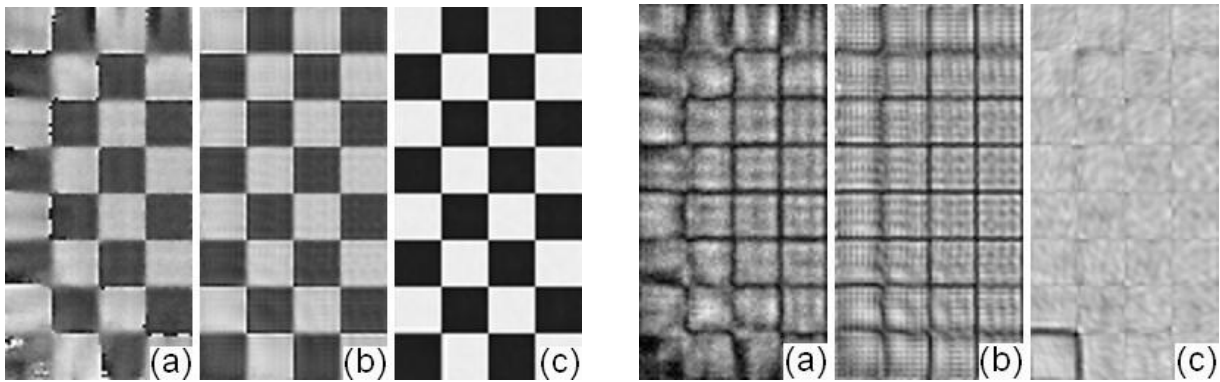

Fig. 1: Fragments of the reconstructed phases (left image) and amplitude (right image) obtained by (a) SBMIR, RMSE($\mathbf{\varphi}_0$)=0.58, RMSE($\mathbf{a}_0$)=0.35; (b) AL, RMSE($\mathbf{\varphi}_0$)=0.26, RMSE($\mathbf{a}_0$)=0.23 and (c) D-AL, RMSE($\mathbf{\varphi}_0$)=0.036, RMSE($\mathbf{a}_0$)=0.026.

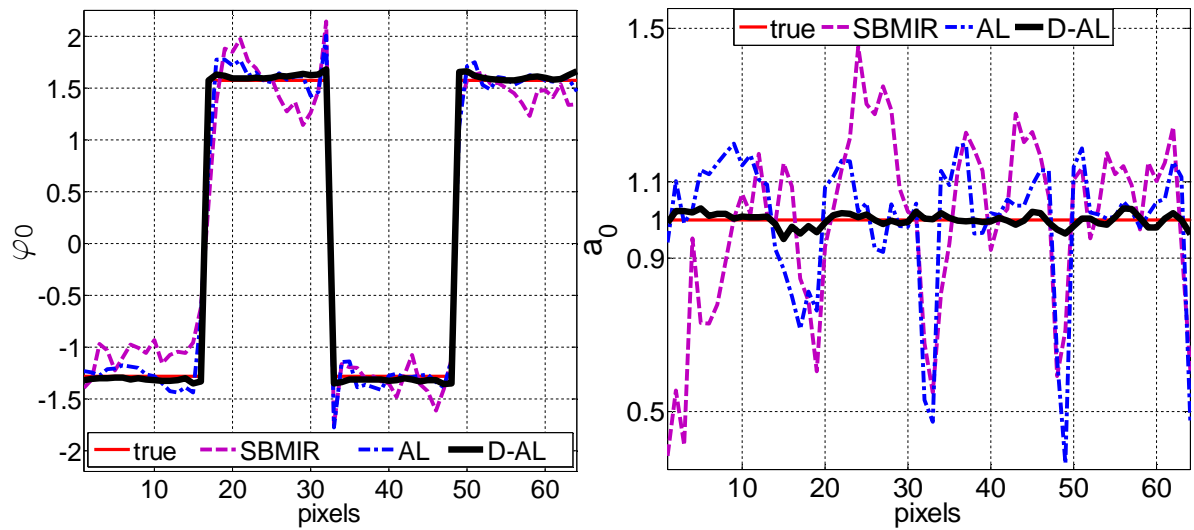

Fig. 2: Cross-sections of the reconstructed object phase (left image) and amplitude (right image), for the test presented in Fig. 1.

**Conclusions**

The developed D-AL algorithm is a further development of the recent AL [7] with the additional BM3D filtering of the object phase and amplitude. It is shown, that this filtering dramatically improves the reconstruction accuracy and imaging. The Matlab codes of the D-AL algorithm used for numerical simulations and more simulation materials are available on our web page http://www.cs.tut.fi/~lasip/DDT/